\documentclass{article}

\usepackage{arxiv}

\usepackage{graphics} %
\usepackage{epsfig} %
\usepackage{amsmath} %
\usepackage[ruled]{algorithm2e}
\SetKwInput{KwInput}{Input}                %
\SetKwInput{KwOutput}{Output}
\usepackage{subfig}
\usepackage{siunitx}
\usepackage{url}
\usepackage{hyperref}       %
\title{\LARGE \bf
Behavioral decision-making for urban autonomous driving in the
presence of pedestrians using Deep Recurrent Q-Network
}

\renewcommand{\today}{\vspace{-1em}}

\author{
	Niranjan Deshpande\\
    Univ. Grenoble Alpes, Inria, 38000 Grenoble, France\\
	\texttt{Niranjan.Deshpande@inria.fr} 
	\And
	Dominique Vaufreydaz\\
    Univ. Grenoble Alpes, CNRS, Grenoble INP, LIG, 38000 Grenoble, France\\
	\texttt{Dominique.Vaufreydaz@univ-grenoble-alpes.fr}
	\And
	Anne.Spalanzani@inria.fr\\
    Univ. Grenoble Alpes, Inria, Grenoble INP, 38000 Grenoble, France\\
	\texttt{Anne.Spalazani@inria.fr}
}

\hypersetup{
pdftitle={Behavioral decision-making for urban autonomous driving in the presence of pedestrians using Deep Recurrent Q-Network},
pdfsubject={},
pdfauthor={Niranjan Deshpande, Dominique Vaufreydaz, and Anne Spalazani},
pdfkeywords={Deep Recurrent Q-Network, Autonomous vehicle},
}

\begin{document}

\maketitle
\thispagestyle{empty}
\pagestyle{empty}

\begin{abstract}
Decision making for autonomous driving in urban environments is
challenging due to the complexity of the road structure and the
uncertainty in the behavior of diverse road users. 
Traditional methods consist of manually designed
rules as the driving policy, which require expert domain knowledge,
are difficult to generalize and might give sub-optimal results as the environment gets complex. Whereas, using reinforcement learning, optimal driving policy could be learned and improved automatically through several interactions with the environment. However, current research in the field of reinforcement learning for autonomous driving is mainly focused on highway setup with little to no emphasis on urban
environments. In this work, a deep reinforcement learning based
decision-making approach for high-level driving behavior is proposed
for urban environments in the presence of pedestrians. For this, the
use of Deep Recurrent Q-Network (DRQN) is explored, a method
combining state-of-the art Deep Q-Network (DQN) with a long term
short term memory (LSTM) layer helping the agent gain a memory of the
environment. A 3-D state representation is designed as the input
combined with a well defined reward function to train the agent
for learning an appropriate behavior policy in a real-world like urban
simulator. The proposed method is evaluated for dense urban scenarios
and compared with a rule-based approach and results show that the
proposed DRQN based driving behavior decision maker outperforms
the rule-based approach.

\end{abstract}


\section{INTRODUCTION}
An autonomous driving system consists of two main sub-systems, namely: a perception sub-system and a navigation sub-system. The navigation  sub-system  is  responsible  for  moving  the  autonomous  vehicle from a source point \textit{A} in the environment to a destination point \textit{B}. For this, the navigation sub-system usually  uses  a decision making module,  a planning and control module  and  an actuator mechanism.  The task of decision making is to make an appropriate driving decision according to the information received from the perception sub-system about the surrounding environment and then a plan  for  drivable path is generated by a path planner which is further passed to the control module.  Reasonable decision-making  in  a  variety  of  complex  environments  is  a  great  challenge,  and  it  is  impossible to enumerate coping strategies in various situations.  Hence, a method that can learn a suitable behavior from its own experiences would be desirable.

Reinforcement learning (RL), a paradigm of machine learning, learns a policy automatically without expert domain knowledge. This is done by simulating various types of different scenarios, even some dangerous ones. 
Deep learning (DL) combined with reinforcement learning (RL) have led to the emergence of deep reinforcement learning (deep RL) based approaches, with Deep Q-Networks (DQN) being one of most commonly used deep RL algorithm. It has demonstrated good results in many areas such as video games \cite{c1}, control theory \cite{c2} and robotics \cite{c3}. As a result, many deep RL based methods are being presented for autonomous driving as well. However, most of these approaches focus on highway driving scenarios and with limited to no focus on complex urban scenarios.

This work focuses on the problem of autonomous driving in urban environments with unsignalized  intersections and in the presence of pedestrians.
A deep RL agent is designed for high-level behavioral decision making combined with a classical PID controller for low-level actuation.
The agent is trained in a simulator to learn the desired driving policy. This proposed approach is compared with a rule-based one for performance evaluation.

The main contributions of this work are as follows:
\begin{itemize}
    \item  To learn high-level behavioural decision making,
    a Deep Recurrent Q-Network (DRQN) \cite{c4} is developed. The DRQN uses a Long Short-Term Memory (LSTM) layer at the end of a typical Deep Q-Network (DQN) \cite{c1} giving memory to deep RL agent.
    \item A unique multi-channeled image like representation is used to train the deep RL agent. This representation can store a variety of information (geometry as well as semantic) necessary for training in complex urban environments.
    \item Instead of assuming an ideal low level controller, this work combines a more realistic PID control for longitudinal actuation.
    For this, appropriate information about the agents state is passed to the network so that the decision maker learns the desired policy taking into account the behavior of the PID controller.
\end{itemize}

The remainder of this paper is structured as follows: A description of deep reinforcement learning algorithms is mentioned in Section \ref{background} and related work is discussed in Section \ref{related_work}. Section \ref{problem_formulation} then introduces the complete framework and the proposed deep RL based behavioral decision maker method.
In Section \ref{experiments}, details about the training process, the simulation setup and performance evaluation results are mentioned. 
The final section is assigned to concluding remarks and future work discussion.
\section{BACKGROUND}\label{background}
\subsection{Reinforcement Learning}
In reinforcement learning (RL), at each time step \textit{t}, the agent receives an observation of the current state $s_{t}$ of the environment and then it decides to take an action $a_t$ based on a policy $\pi$. 
After the action is performed in the environment, the agent receives a numerical reward $r_{t}$ and the next state $s_{t+1}$ of the environment.
The goal of an RL agent is to learn an optimal policy, which maximizes the sum of future discounted rewards $R$, defined as :
$$ R = \sum_{t=0}^{T} \gamma \cdot r_{t}$$
where $\gamma$ is a discount factor defining the importance of future rewards. 

\subsection{Q-Learning}
In Q-learning, a type of RL algorithm, the agent tries to learn a state-action value function, also called the Q-function. 
This function for a policy $\pi$ is defined as the expected return for executing an action $a$ in a state $s$:
\begin{equation}
    Q^{\pi}(s, a) = E [ \ R_{t}|s_{t} = s, a_{t} = a] \
\end{equation}
The function gives a Q-value for each state-action pair indicating how good taking an action might be at a particular state.
The goal of Q-learning algorithm is to find an optimal value function $Q^{*}(s, a)$ defined as:
\begin{equation}
    Q^{*}(s, a) = \underset{\pi}{max} E [ \ R_{t}|s_{t} = s, a_{t} = a] \ = \underset{\pi}{max} Q^{\pi}(s, a)
\end{equation}

\subsection{Deep Q-Networks}
In Deep Q-Networks (DQN) \cite{c1}, a multi-layered neural network, with parameters $\theta$, is used to approximate the value function and is denoted as $Q(s, a; \theta)$.
The DQN outputs a vector of action values, given a state $s$.

Two important ingredients of DQN are \cite{c22}: experience replay and a target network.
Replay memory improves the performance of learning by eliminating the correlations in the observation sequences.
The target network, with parameters $\theta^{-}$, is same as the main network, however, its parameters are updated from the main network after every $t$ steps, that is, $\theta^{-}_{t} = \theta_{t}$. The goal is to minimize the average of the loss:
\begin{equation}
 L(\theta_{t}) = \Big( \underbrace{(r + \gamma\underset{a}{max}Q{}(s_{t+1}{}, a;\theta^{-}_{t})}_\text{\normalfont $y$} -\ Q(s,a;\theta^{-}_{t}) \Big)^2 \label{eq:1}
\end{equation}
where $t$ is the time step and $y$ is the update target used by DQN.

\subsection{Double Deep Q-Networks}\label{ddqn}
Conventional DQN faces the problem of overestimation.
The max operation in DQN uses the same Q-function network for both, action selection and evaluation, causing overestimation of action values.
This is solved by using Double Deep Q-Network (DDQN) which uses the main network for
evaluating the greedy policy, while the target network is used to estimate the value of this policy. 
The target for loss function in DDQN, is defined as:
\begin{equation}
    y^{DDQN} = r + \gamma Q{}(s_{t+1}{}, \underset{a}{argmax}Q(s_{t+1},a;\theta_{t});\theta^{-}_{t})
\end{equation}
In the $argmax$ term, the main network with weights $\theta_{t}$, is used 
for action selection, while the target network with weights $\theta^{-}_{t}$ is for action evaluation.

\subsection{Long Short-Term Memory (LSTM)}
A regular recurrent neural network has vanishing gradients problem while dealing with longer sequences.
This problem is solved by using an LSTM \cite{c23} which stores information in memory cells, carefully regulated by using input, output and forget gates.
These gates can learn to decide if a memory cell is is to be kept or removed.
This helps in storing both recent and older relevant observations to be utilized by the network.

\section{RELATED WORK}\label{related_work}
Given the recent developments in deep learning, interest in using learning based approaches for autonomous driving has increased significantly. 
There has also been immense interest in using deep RL for autonomous navigation given its ability to self learn from experiences. A detailed survey of these approaches could be found in \cite{c8, c9}.

One of the important aspect for deep RL is way in which the state is represented and therefore, many different state representations are being used. In \cite{c10}, \cite{c11} raw RGB images are used as a state representation for the RL agent. 
Another common practice for state representation is the use of a one-dimensional vector storing relative position, speed and lane of the surrounding vehicles. For instance, \cite{c12} uses information of 8 surrounding vehicles in the form of a vector for the problem of speed and lane change decision making on highway environment. An image-like grid (matrix) format for state representation has been used in some approaches \cite{c13}, \cite{c15} for dealing with intersection crossing scenarios. In \cite{c13} grid cells are used for representing the speed of surrounding vehicles for lane change decision making. The traffic information is extracted from the simulator to create a lane based grid representation.
In \cite{c15}, a multi-layered grid representation is used for urban traffic light control.
It consists of two matrices of size ${64\times64}$, one for surrounding vehicle position and another for their speeds.

Raw RGB images contain very high dimensional data such as textures and appearance of roads and objects. Hence for generalization, large data set is required covering each of these dimensions. The vector based representation works well for highway environments where the road geometry and the traffic is structured and the agents behavior is affected by leading and preceding vehicles in the same and opposite lanes.
However, urban environments have complex road structures such as intersections and crossings. It consists of different vulnerable road users (VRUs) such as pedestrians and bicyclists with motions that are  less  constraint  in  comparison  to  the vehicles. 
Hence, in this work an image like multi-layered grid representation state representation is employed due to its capability to represent a variety of features essential for decision making in urban environment context.

One problem when using reinforcement learning for autonomous driving is the assumption that 
the environment is fully observable, that is, there is complete information about the state of the environment at any step \cite{c16}. This problem is solved by some authors, for example, in \cite{c17}, an approach called Deep Recurrent Q-Network (DRQN) is developed which uses a Long Short-Term memory (LSTM) layer to estimate the Q-function. However, these approaches use feature vector state representation for the DRQN decision maker. Hence, this work takes inspirations from \cite{c4} which uses a similar DRQN approach to train an agent play Atari 2600 games with partial observability. However, in \cite{c4} the input to the agent is a multi-channel raw image which passes through several convectional neural layers and the output of these layers is then passed to an LSTM layer which selecting appropriate action.

\section{PROBLEM FORMULATION}\label{problem_formulation}
Autonomous driving in urban environments is challenging due to the complexity of the road structure and the uncertainty in the behavior of diverse road users.
Interpreting the intentions and motions of other road users becomes important, especially for vulnerable road users (VRUs) such as pedestrians as their motions  are  less  constraint  in  comparison  to  vehicles  and  even  slightest  collision  is  likely  to  be fatal.  Often,  pedestrians perform jaw-walking putting themselves into a vulnerable situation in the context of autonomous driving.

This work focuses on the high-level behavioural decision making aspect of autonomous driving in the presence of pedestrians for a crowded urban environment with unsignalized intersections.
A framework is proposed to reduce the complexity of the task by dividing it into sub-problems: high-level decision making and low-level control. For decision making, a  Deep Recurrent Q-Network (DRQN) is used and combined with a PID longitudinal controller.
The objective is to autonomously drive along an urban road which has unsignalized intersections with crosswalks and control the velocity such that collision with pedestrians could be avoided in a comfortable manner.

\subsection{System architecture}
\begin{figure}[h]
     \centering
     \includegraphics[width=\linewidth]{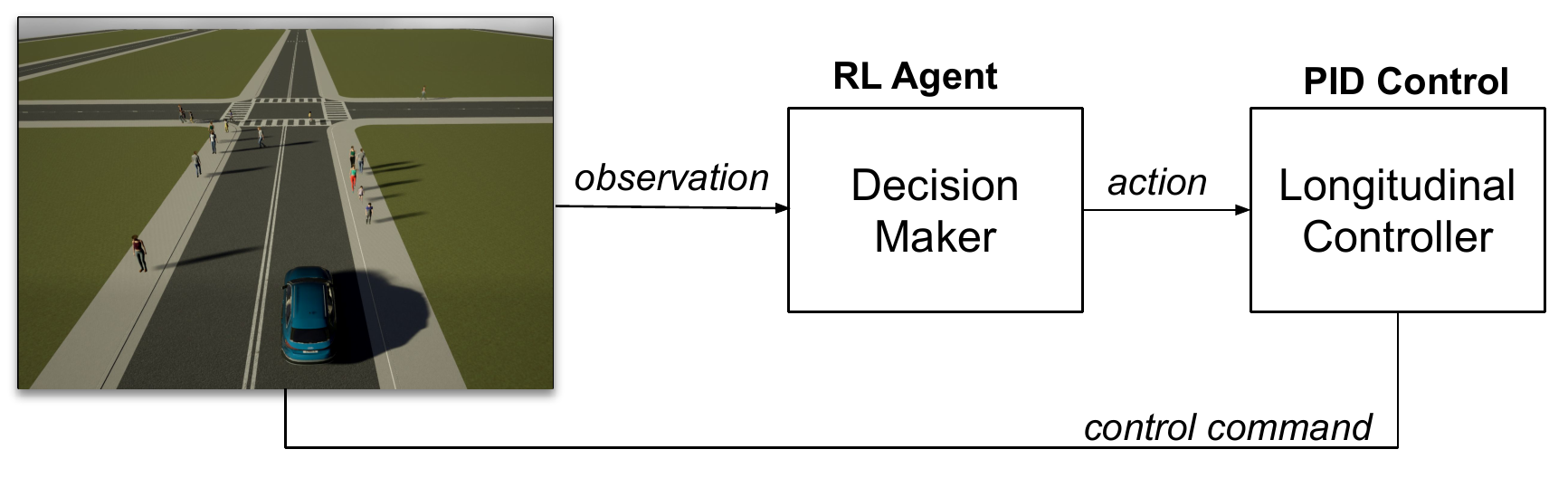}
     \caption{Framework architecture.}
     \label{fig:framework}
\end{figure}

Fig. \ref{fig:framework} shows the proposed framework architecture.
An urban environment is defined as the world around the ego vehicle, including all pedestrians of interest.  
The environment is developed in a simulator explained in Section \ref{simulation} and sends observations to the agent at each instance. Based on the learned decision making policy, the agent chooses appropriate high level actions (details in Subsec. \ref{action}).
These actions are then sent to a longitudinal PID controller which calculates appropriate throttle values given to the ego vehicle.

\subsection{POMDP formulation}
This work formulates the autonomous driving decision making in urban environment as a Partially Observable Markov Decision Process (POMDP), as the ego-vehicle cannot observe the intentions, that is, the intended destination of the surrounding pedestrians.
The input state representation, action space and the reward function for learning the desired driving policy are mentioned below:
\hfill\\
\subsubsection{State space}
\hfill\\
\indent
The representation for state space needs to have enough and useful information about the environment and the agent itself for urban autonomous driving.
Hence, a multi-layered (3-D) grid representation, derived from our previous work \cite{c21} is used.
It resembles an image-like representation with multiple channels, with each channel corresponding to a particular layer.
To generate this grid, a bird's eye view of environment is discretized into a grid in Cartesian coordinates. 
The resolution of each grid is of size ${1m\times1m}$, assuming the footprint of a pedestrian is approximately ${1\, meter}$.

The first layer is a position matrix, where a grid value is set to 1 indicating presence of pedestrian
and absence of pedestrian is indicated by 0 value.
The second layer stores relative heading information about the corresponding pedestrian, and the third layer is use for relative speed. The fourth layer stores semantic information in the form of type of road structure occupied by the pedestrian, namely: walkway, crossing or road. 
\hfill\\
\subsubsection{Action Space}\label{action}
\hfill\\
\indent 
The goal is to learn an optimal decision making policy to select appropriate high-level driving behavior.
For this 4 actions are defined, each corresponding to a specific behavior, namely:
\begin{itemize}
    \item \textit{\textbf{accelerate:}} to speed up by gradually increasing the current speed by $+1\, km/hr$.
    \item \textit{\textbf{slow down:}} to slow down by gradually decreasing agents current speed by $-1\, km/hr$.
    \item \textit{\textbf{brake:}} to apply brakes and stop.
    \item \textit{\textbf{steer:}} to advance by maintaining the current speed.
\end{itemize}

\begin{figure}[h]
    \centering
    \includegraphics[width=\linewidth]{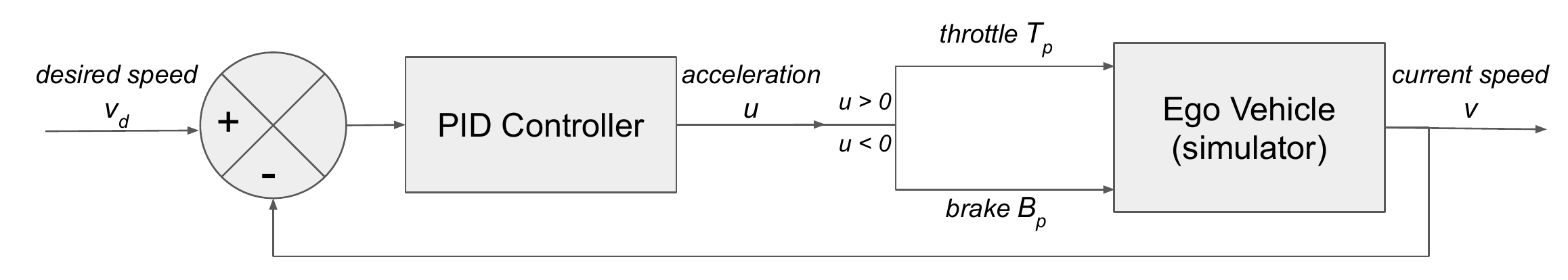}
    \caption{Longitudinal PID Controller \cite{c24}}
    \label{fig:pid}
\end{figure}
The selected action (behavior) is then passed to a PID controller, shown in Fig. \ref{fig:pid}, which in turn generates appropriate throttle and brake values.
This is explained in Algo. \ref{alg:the_alg}. Basically, each action updates the desired speed ($v_{d}$) input to the controller.
Then based on the error between current ego vehicle speed ($v$) and the updated desired speed ($v_{d}$), the controller generates appropriate acceleration ($u$) using the equation:
\begin{equation}
    u = K_{p}(v_{d} - v) + \int_{0}^t (v_{d} - v)dt + \frac{d(v_{d} - v)}{dt}
\end{equation}
The acceleration value further decides the throttle position $T_{p}$ and braking $B_{p}$,
using the following conditions:
\begin{equation}
    T_{p} =
    \begin{cases}
      u, & \textit{if}\ u\geq1 \\
      0, & \textit{if}\ u<0
    \end{cases}
    \hspace{0.2cm},\hspace{0.3cm}
    B_{p} =
    \begin{cases}
      0, & \textit{if}\ u\geq1 \\
      -u, & \textit{if}\ u<0
    \end{cases}
\end{equation}
\begin{algorithm}
\caption{High-level action to throttle, brake conversion}
\label{alg:the_alg}
\setlength{\algomargin}{-20em}
\KwInput{\textit{Actions} = \{\textit{accelerate}, \textit{slow\_down}, \textit{brake}, \textit{steer\}}}
\KwOutput{$Throttle(T_{p}$), $Brake(B_{p})$\\}
$T_{p}, B_{p} \leftarrow 0$	 \\
{\SetAlgoNoLine%
    \If{acceleration}
    {
        $v_{d} \leftarrow + 1 $  \\
        $T_{p}, B_{p} \leftarrow compute\_throttle\_brake(v_{d})$ \\
    }
    \ElseIf{slow\_down}
    {
        $v_{d} \leftarrow - 1 $  \\
        $T_{p}, B_{p} \leftarrow compute\_throttle\_brake(v_{d})$ \\
    }
    \ElseIf{steer}
    {
        $v_{d} \leftarrow +0 $  \\
        $T_{p}, B_{p} \leftarrow compute\_throttle\_brake(v_{d})$ \\
    }
    \ElseIf{brake}
    {
        $T_{p} \leftarrow 0$ \\
        $B_{p} \leftarrow 1$\\
    }
    \KwRet $T_{p}, B_{p}$ 
    }
\end{algorithm}

\subsubsection{Reward Function}
\hfill\\
\indent
A reward function helps an  agent in learning its desired policy.
In the context of autonomous driving in urban environment, two main aspects are considered while developing the reward function: safety and efficiency.
With respect to safety, an autonomous vehicle should be able to avoid any collisions while driving. It also must do this without causing discomfort to the pedestrians and its passengers.
For efficiency, the autonomous vehicle should drive as fast as possible but without exceeding the speed limits.
After evaluating several reward combinations, the following function is used to formulate the reward:
  \begin{equation}
    r =
    \begin{cases}
      r_{c} & \textit{for collision}\\
      r_{nc} & \textit{for near collision}\\
      r_{v} & \textit{for normal driving}
    \end{cases}
  \end{equation}
where $r_{c}= -10$, is a large penalty if there is a collision.
A small penalty of $r_{nc}$ is also given in case of near collision situations, to discourage the agent from driving too close to the pedestrians and causing them discomfort. To compute $r_{nc}$, a time-to-collision (TTC) based piece-wise function is used. Pedestrians with $TTC <= 3$ seconds are considered and the $r_{nc}$ is updated using the following equation:
  \begin{equation}
    r_{nc} = TTC - 3.0
  \end{equation}
  
\vspace{2cm}
Hence, lesser the time to collision, higher will be the penalty $r_{nc}$.
The term $r_{v}$ represents a reward for encouraging the agent to maintain a higher speed within the traffic speed limits. The reward for speed is computed using the following function:
  \begin{equation}
    r_{v} =
    \begin{cases}
      1 - \lambda\cdot(v_{ref} - v_{ev}), & \textit{if} \quad \text{0} < v_{ev} \leq v_{ref}\\
      -1, & \textit{if}\quad{ v_{ev}} \leq \text{0}\\
      -0.5, & \textit{if}\quad v_{ev} > v_{ref}\\
    \end{cases}
  \end{equation}
where $\lambda = 1/v_{ref}$, $v_{ev}$ is ego vehicle speed and $v_{ref}$ is the maximum allowed speed.
A penalty of $-1$ is assigned to the ego vehicle to encourage it to keep moving and a penalty of $-0.5$ is applied for exceeding the speed limit.\\
\section{EXPERIMENTS AND RESULTS}\label{experiments}
\subsection{Neural Network Model}
\vspace{-5.0mm}
\begin{figure}[ht]
    \centering
    \includegraphics[scale=0.82]{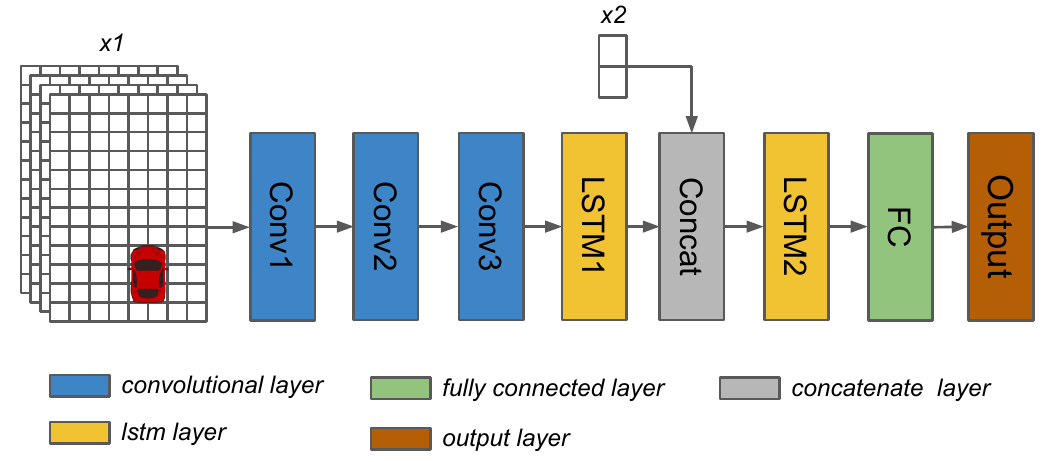}
    \caption{The network architecture}
    \label{fig:nn}
\end{figure}
The network architecture used in this work consists of stacked LSTM structure inspired from \cite{c20} and is shown in Fig. \ref{fig:nn} . %
The input $x_{1}$ to the network is a multi-layered grid of size ${45\times30\times4}$.
This input is propagated through three convolutional layers with each convolutional layer followed by a  ReLU activation function.
The third convolutional layer is then flattened and is fed into an LSTM layer.
The output of this layer is then input to the following LSTM layer along with input $x_{2}$ which consists of ego vehicle's speed and the selected action.
Finally the output of the stacked LSTM is fed to the output layer through a fully connected layer.
The parameters of the network are mentioned in the Tab. \ref {table_nn} below:
\begin{table}[ht]
\begin{center}
\begin{tabular}{|c|c|c|c|c|c|}
\hline
\textbf{Layer} & {Activation} & {Kernel size} & {Stride} & {Filters} & {Units}\\
\hline
Conv1 & ReLU & 8*6 & 4 & 32 & -\\
Conv2 & ReLU & 4*3 & 3 & 64 & -\\
Conv3 & ReLU & 2*2 & 2 & 64 & -\\
\hline
LSTM1 & - & - & - & - & 256\\
LSTM2 & - & - & - & - & 256 \\
\hline
FC  & ReLU & - & - & & 256 \\
\hline
\end{tabular}
\end{center}
\caption{Parameters of the neural network}
\label{table_nn}
\vspace{-4mm}
\end{table}
\subsection{Training}
During training, the agent interacts with the environment following an $\epsilon$-greedy policy mentioned in section \ref{exploration_strategy}.
The network is trained using Double Deep Q-Network (DDQN) algorithm, descried in Sec. \ref{ddqn}, which consists of one main Q network and a target Q network with the same architecture.
The parameters of the main network are updated using an Adam optimizer \cite{c18}.
The parameters of the target network are updated periodically after every  $N$ steps, by cloning the main network parameters.
For stability, error clipping is employed by limiting the error term of Eq. \eqref{eq:1} to [-1, 1].
Since the network uses a recurrent layer, sequential updates are required.
The experiences gathered during training are stored as episodes in a replay memory. During training a mini-batch of these episodes is selected uniformly, with each episode consisting of 8 time steps.
After performing several tests, a relatively good set of hyper-parameters have been selected and are mentioned in the Tab. \ref{table_hyper}.

\begin{table}[ht]
\begin{center}
\begin{tabular}{|l|l|l|}
\hline
\textbf{Hyper-parameters} & {Value} & {Description}\\
\hline
Learning rate, $\alpha$ & 0.001 & Learning rate used by Adam \\
Discount factor, $\gamma$  & 0.9 & Discounted future reward\\
Mini-batch size, $M_{mini}$ & 32 & Number of training episodes\\
Target update, $N$ & 10000 & Target network update steps\\
Replay memory size, $M$ & 50 & Max. episodes stored in memory\\
Training episodes, $E$ & 200 & Total episodes used for training\\
Episode steps, $S$ & 1000 & Maximum steps per episode\\
\hline
\end{tabular}
\end{center}
\caption{Hyper parameters of the DDQN agent}
\label{table_hyper}
\end{table}

\subsection{Exploration Strategy} \label{exploration_strategy}
It is a common practice to use an $\epsilon$-greedy policy for Q-learning.
Generally, epsilon greedy approach randomly selects an action from a uniform distribution. 
A random action is selected with probability $\epsilon$ and current optimal action is selected with probability of $1 - \epsilon$. When $\epsilon = 1$, completely random action is selected, whereas, when $\epsilon = 0$ current optimal action is selected. 
In this work, the value of $\epsilon$ is linearly decreased from $\epsilon_{start} = 1.0$ to $\epsilon_{end} = 0.1$ over the training episodes.
However, instead of a uniform distribution, this work employs a modification to improve the learning process. For initial few episodes, the \textit{accelerate} and \textit{steer} actions  are selected with slightly higher probability, to encourage the ego vehicle to learn moving at an early stage and subsequently encounter collision situations sooner.

\subsection{Simulation setup and scenario}\label{simulation}

The proposed method is trained and evaluated in CARLA simulator \cite{c19}, an open source simulator for autonomous driving.
An urban environment with unsignalized intersection is developed in the simulator.
The simulator provides environment information and receives throttle and brake signals for the ego vehicle.
The simulator time step is set at $t =\, $\SI{0.1}{\second}.

A typical urban environment scenario is considered as shown in Fig. \ref{fig:carla}.
At the beginning of each episode, an ego vehicle is spawned at the start position
and follows a list of way-points generated by the simulator to reach its destination.
The task for the ego vehicle is to navigate as fast as possible, within the speed limits, through the way-points to reach its destination while comfortably avoiding collisions with the surrounding pedestrians.
There are 10 pedestrians initially sampled randomly in the vicinity of the ego vehicle, with random desired speed ranging between $0.5-1.5\,km/hr$ and a randomly assigned goal.
Each pedestrian follows one of the three behaviors: legal crossing on crosswalks, illegal jaw walking on roads or normal walking on sidewalks (no crossing) with the probabilities of 0.6, 0.2, 0.2 respectively.
As the ego vehicle moves through its route, pedestrians moving farther away are removed and corresponding number of new pedestrians are sampled in the vicinity of the vehicle, making sure there is always quite busy pedestrian traffic. 

\subsection{Results}

\begin{figure}
    \begin{minipage}[b]{0.5\textwidth}
        \centering
        \includegraphics[scale=0.3]{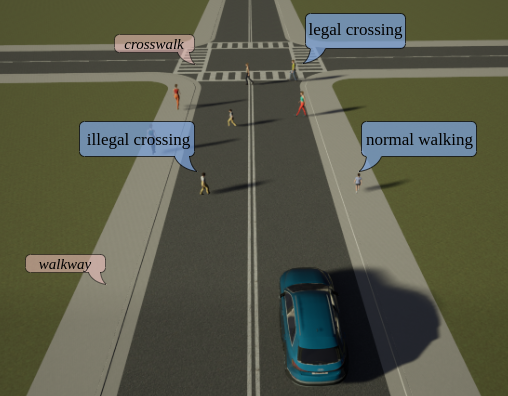}
        \caption{Simulation environment and training scenario}
        \label{fig:carla}
    \end{minipage}
    \begin{minipage}[b]{0.5\textwidth}
        \centering
        \subfloat[]{\includegraphics[width=0.48\textwidth, height=4cm]{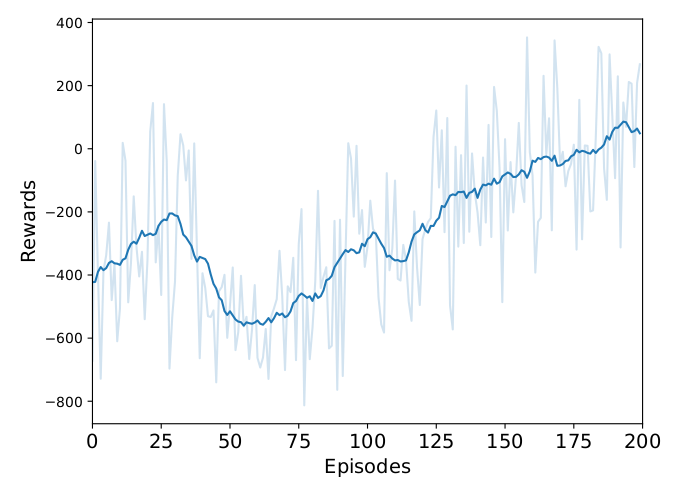}} 
        \subfloat[]{\includegraphics[width=0.48\textwidth, height=4cm]{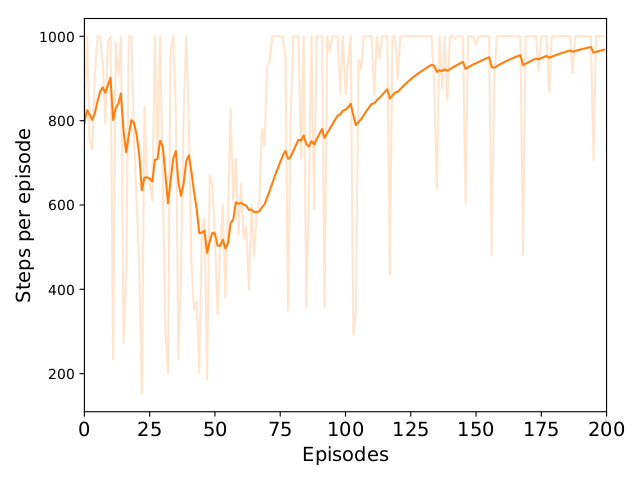}} 
        \caption{Training performance}
        \label{fig:graphs}
    \end{minipage}
\end{figure}

The proposed DRQN agent is trained for 200 episodes. For each episode the maximum step limit is set to 1000 steps, however, an episode terminates if the goal is reached or if there is a collision. 
The average reward and the number of steps executed per episode are shown in Fig. \ref{fig:graphs}.
As seen in the reward graph, in the beginning, the reward gradually increases as the agent learns to move. However, there is a sudden drop in the reward, because, as the agent learns to move it starts encountering collisions and getting penalties. After some episodes, the agent starts learning the desired behavior and hence the reward starts increasing again.
Similarly, initially the number of steps per episode are high as the agent is learning to move. After some episodes, there is a drop in number of steps due to collisions and as training progresses further the number of steps start increasing again.

The performance of the trained agent is evaluated by comparing it with a rule-based method of CARLA simulator.
This method always tries to steer the ego vehicle with maximum speed to reach the target destination while avoiding collisions with other vehicles and pedestrians.
This is done by applying emergency braking in case of a collision situation.
The maximum speed ($v_{max}$) is set to $15\,km/hr$, by fixing the desired speed of the PID controller to $v_{d} = v_{max}$.
The rule-based agent follows a conservative policy, wherein, the ego-vehicle starts applying braking for legal and illegal crossing pedestrians which are in the vicinity of upto $7\, meters$, in front of the vehicle. 

Both the methods are tested for 10 episodes each and are evaluated using the metrics of percentage of collision free episodes, average speed (in $km/hr$) and average distance travelled (in $meters$).
The results are depicted in Tab. \ref{tab:compare}.
In terms of safety, the proposed method performs much better than a standard rule-based one. Also, the average distance travelled without any collisions is much better with the proposed method.
In terms of average speed, the rule-based method performs better.
This is because, since the DRQN has memory, it anticipates the behavior of nearby pedestrians and slows down in case of a possible collision like situations.
The collision and speed performance of the proposed DRQN method could be further enhanced by training the agent on more episodes.
\vspace{2 mm}
\begin{table}[ht]
  \begin{center}
    \begin{tabular}{|l|c|c|c|} %
       \hline
       & \textbf{Collision free} & \textbf{Average} & \textbf{Distance}\\
       & \textbf{episodes} & \textbf{speed} & \textbf{travelled}\\
      \hline
      rule-based agent & $40\%$ & 7.79 & 82.6\\
      DQN-based agent & $70\%$ & 6.09 & 123.1\\
      \hline
    \end{tabular}
  \end{center}
    \caption{Comparison between rule-based and DQN approaches}
    \label{tab:compare}
\end{table}
\section{CONCLUSION}\label{conclusion}
\vspace{-0mm}
In this paper, a method is proposed to handle the problem of high-level behavioral decision making in autonomous driving amongst pedestrians.
For that an urban environment with unsignalized intersections is developed in a real-world-like simulation environment. 
A Deep Recurrent Q-Network (DRQN) based agent is used for decision making combined with a low level PID controller.
Given a multi-layered state representation and a well designed reward function, the trained agent is able to take appropriate actions amongst pedestrians.
The trained agent is compared with a rule-based method and the results demonstrate the effectiveness of the proposed method.
Although the proposed method is better than rule-based, 
its performance could further be improved. 
In the future, the agent will be trained on a larger dataset using more computational resources to push the limit of the proposed method and also to further improve the learning efficiency. 
More advanced deep RL algorithms could be explored for behavioral decision making for urban autonomous driving in the future.
Another potential future work is to consider more complex urban scenarios by incorporating vehicles as well, along with pedestrians. 
\section{ACKNOWLEDGMENT}
This work was funded under project CAMPUS (Connected Automated Mobilty Platform for Urban Sustainability) sponsored by Programme d'Investissements d'Avenir (PIA) of french Agence de l'Environnement et de la Ma\^itrise de l'\'Energie (ADEME).

\addtolength{\textheight}{-12cm}   %

\end{document}